%% file: main.tex
\documentclass[runningheads]{llncs}
\usepackage{xcolor}
\usepackage{booktabs}
\usepackage{epsfig}
\usepackage{graphicx}
\usepackage{amsmath}
\usepackage{amssymb}
\usepackage{siunitx}
\usepackage{graphicx}
\usepackage{subcaption}
\usepackage[pagebackref=true,breaklinks=true,letterpaper=true,colorlinks,bookmarks=false]{hyperref}
\usepackage{pdfpages}

\usepackage{tikz}
\usepackage{comment}
\usepackage{amsmath,amssymb} % define this before the line numbering.
\usepackage{color}

\newcommand{\pose}{\boldsymbol{\theta}}
\newcommand{\shape}{\boldsymbol{\beta}}
\definecolor{mariocolor}{rgb}{1,0,1}
\definecolor{hosnacolor}{rgb}{1,0,0}
\definecolor{gerardcolor}{rgb}{0,0,1}

\DeclareMathOperator{\sign}{sign}
\DeclareMathOperator{\clip}{clip}

\begin{document}

%%%%%%%%% TITLE
\title{Body Shape Privacy in Images: Understanding Privacy and Preventing Automatic Shape Extraction}
\titlerunning{Body Shape Privacy in Images}

\pagestyle{headings}
\mainmatter 
\author{Hosnieh Sattar\inst{1}\and
Katharina Krombholz\inst{2}\and
Gerard Pons-Moll\inst{1}\and
Mario Fritz\inst{2}}
\authorrunning{H. Sattar et al.}
\institute{Max Planck Institute for Informatics,
\email{\{sattar,gpons\}@mpi-inf.mpg.de}\\ \and
CISPA Helmholtz Center for Information Security,
\email{\{fritz,krombholz\}@cispa.saarland}\\ 
Saarland Informatics Campus, Saarbr\"ucken, Germany.}

\maketitle
\input{abstract.tex}
\input{intro.tex}
\input{relatedworks.tex}

\input{userstudy.tex}

\input{method.tex}
\input{experiment.tex}

{\small
\bibliographystyle{ieee}
\bibliography{egbib}
}
\input{supp.tex}
\includepdf[pages=-]{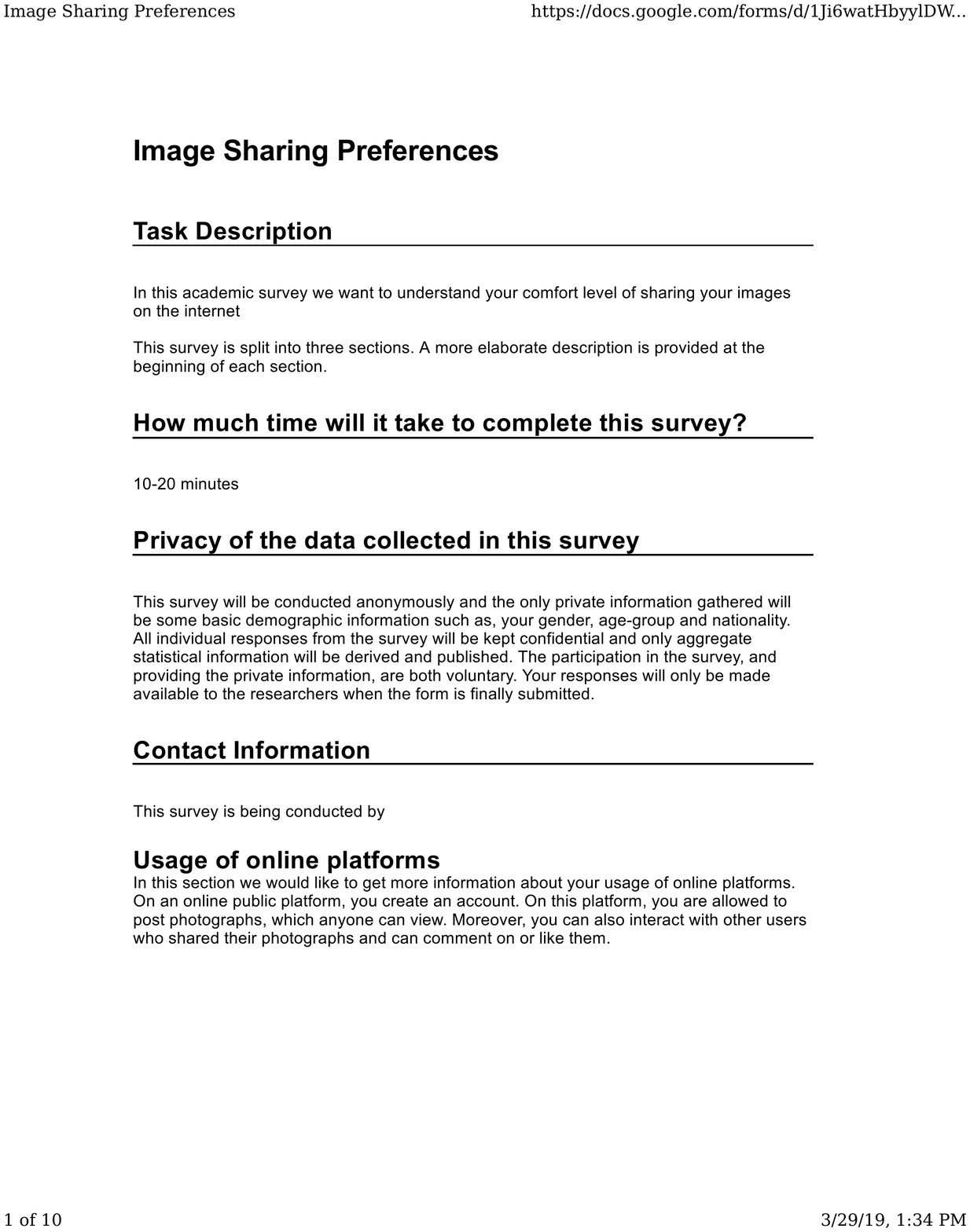} 

\end{document}

%% file: abstract.tex
\begin{abstract}
Modern approaches to pose and body shape estimation have recently achieved strong performance even under challenging real-world conditions. Even from a single image of a clothed person, a realistic looking body shape can be inferred that captures a users' weight group and body shape type well. This opens up a whole spectrum of applications -- in particular in fashion -- where virtual try-on and recommendation systems can make use of these new and automatized cues. However, a realistic depiction of the undressed body is regarded highly private and therefore might not be consented by most people. Hence, we ask if the automatic extraction of such information can be effectively evaded. While adversarial perturbations have been shown to be effective for manipulating the output of machine learning models -- in particular, end-to-end deep learning approaches -- state of the art shape estimation methods are composed of multiple stages. We perform the first investigation of different strategies that can be used to effectively manipulate the automatic shape estimation while preserving the overall appearance of the original image. 
\end{abstract}

%% file: intro.tex
\section{Introduction}

Since the early attempts to recognize human pose in images \cite{598236,Gavrila1999TheVA}, we have seen a transition to real-world applications where methods operate on challenging real-world conditions in uncontrolled pose and lighting. We have seen more recently progress towards extracting richer representations beyond the pose. Most notably, a full body shape that is represented by a 3D representation or a low dimensional manifold (SMPL) \cite{SMPL:2015}. It has been shown that such representations can be obtained from fully clothed persons -- even in challenging conditions from a single image~\cite{smplify} as well as from web images of a person~\cite{sattar19wacv}.
\begin{figure}[t]
\centering
\includegraphics[width=0.8\linewidth, height=\linewidth,keepaspectratio]{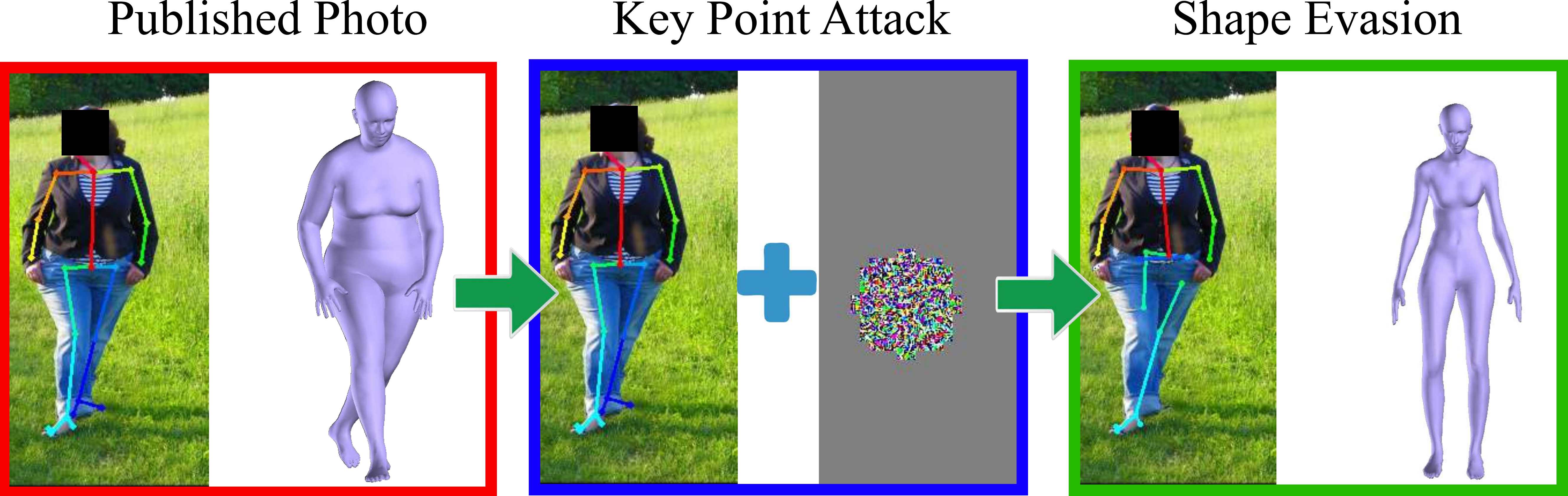}
\caption{A realistic depiction of the undressed body is considered highly private and therefore might not be consented by most people. We prevent automatic extraction of such information by small manipulations of the input image that keep the overall aesthetic of the image. }
\label{fig:comf}
\vspace{-0.2cm}
\end{figure}

On the one hand, this gives rise to various applications -- most importantly in the fashion domain. The more accurate judgment of fit could minimize clothing returns, and avatars and virtual try-on may enable new shopping experiences. Therefore, it is unsurprising that such technology already sees gradual adaption in businesses\footnote{https://www.cnet.com/news/amazon-buys-body-labs-a-3d-body-scanning-tech-startup/}, as well as start-ups \footnote{https://bodylabs.io/en/}.

On the other hand, the automated extraction of such highly personal information from regular, readily available images might equally raise concerns about privacy. Images contain a rich source of implicit information that we are gradually learning to leverage with the advance of image processing techniques. Only recently, the first organized attempts were made to categorize private information in images \cite{orekondy17iccv} to raise awareness and to activate automatic protection mechanisms.

To control private information in images, a range of redaction and sanitization techniques have been introduced \cite{orekondy17connect,sun2018hybrid,edgar2018cscs}. For example, evasion attacks have been used to disable classification routines to avoid extraction of information. Such techniques use adversarial perturbations to throw off a target classifier. It has been shown that such techniques can generalize to related classifiers \cite{oh17iccv}, or can be designed under unknown/black-box models \cite{joon2018iclr,ChenAISEC2017,BrendelARXIV2017a,SuARXV2017,Anonymous2018NIPSb}.

Unfortunately, such techniques are not directly applicable to state-of-the-art shape estimation techniques \cite{smplify,alldieck2018video,Lassner,alldieck19cvpr,habermann2019TOG}, as they are based on multi-stage processing. Typically, deep learning is used to extract person \emph{keypoints}, and a model-fitting/optimization stage leads to the final keypoint estimation of pose and shape. As a consequence, there is no end-to-end architecture that would allow the computation of an image gradient needed for adversarial perturbations.

Today, we are missing successful evasion attacks on shape extraction methods. In this paper, we investigate to what extent shape extraction can be avoided by small manipulations of the input image (\autoref{fig:comf}). We follow the literature on adversarial perturbations and require our changes in the input image to be of a small Euclidean norm. After analyzing a range of synthetic attack strategies that operate at the keypoints level, we experimentally evaluate their effectiveness to throw off multi-stage methods that include a model fitting stage. These attacks turn out to be highly effective while leaving the images visually unchanged. In summary, our contributions are:
\begin{itemize}
\item An orientative user study of concerns w.r.t. privacy and body shape estimation in different application contexts.
\item Analysis of synthetic attacks on 2D keypoint detections.
\item A new localized attack on keypoint feature maps that requires a smaller noise norm for the same effectiveness. 
\item Evaluation of overall effectiveness of different attacks strategies on shape estimation. We show the first successful attacks that offer an increase in privacy with negligible loss in visual quality.
\end{itemize}

%% file: relatedworks.tex
\section{Related Works}
This work relates to 3D human shape estimation methods, privacy, and adversarial image perturbation techniques. We will here cover recent papers in these three domains and some of the key techniques directly relating to our approach.

\textbf{Privacy and Computer Vision.}
Recent developments in computer vision techniques \cite{Deng2009CVPR,Krizhevsky2012NIPS,He2016CVPR,Oh2015ICCV}, increases concerns about extraction of private information from visual data such as age \cite{Bauckhage2010AgeRI}, social relationships \cite{Wang2010SeeingPI}, face detection \cite{Sun2017FaceDU,Viola2001RobustRF}, landmark detection \cite{Zheng2009TourTW}, occupation recognition \cite{shao2013you}, and license plates \cite{Zhou2012PrincipalVW,Zhang2006LearningBasedLP,Chang2004AutomaticLP}. 
Hence several studies on keeping the private content in visual data began only recently such as adversarial perturbations \cite{moosavi2016universal,papernot2016distillation}, automatic redaction of private information \cite{orekondy17connect}, predicting privacy risks in images \cite{orekondy17iccv}, privacy-preserving video capture \cite{aditya2016pic,pittaluga2015privacy,neustaedter2006blur,raval2014markit}, avoiding face detection \cite{wilber2016can,harvey2012cv}, full body re-identification \cite{oh2016faceless} and privacy-sensitive life logging \cite{Hoyle:2015:SLP:2702123.2702183,screenavoider2016chi}.
None of the previous work in this domain studied the users' shape privacy preferences. 
Hence, we present a new challenge in computer vision aimed at preventing automatic 3D shape extraction from images. 

\textbf{3D Body Shape Estimation.}
Recovery of 3D human shape from a 2D image is a very challenging task due to ambiguities such as depth and unknown camera data. This task has been facilitated by the availability of 3D generative body models learned from thousands of scans of people~\cite{anguelov2005scape,pons2015dyna,SMPL:2015}, which capture anthropometric constraints of the population and therefore reduce ambiguities. 
Several works~\cite{sattar19wacv,NIPS2007_3271,5459300,zhou2010parametric,10.1007/978-3-642-15558-1_22,smplify,MuVS:3DV:2017,hasler2010multilinear,zhou2010parametric} leverage these generative models to estimate 3D shape from single or multiple images, using shading cues, silhouettes and appearance. Recent model based approaches are using deep learning based 2D detections~\cite{cao2017realtime} -- by either fitting a model to them at test time~\cite{alldieck19cvpr,sattar19wacv,smplify,alldieck2018video,habermann2019TOG} or by using them to supervise bottom-up 3D shape predictors~\cite{omran2018NBF,Pavlakos_2018_CVPR,Kanazawa_2018_CVPR,tung2017self,tan2017indirect,alldieck19cvpr}. Hence, to evade recent shape estimators, we study different strategies to attack the 2D keypoint detections while preserving the overall appearance of the original image.

\textbf{Adversarial Image Perturbation.}
Adversarial examples for deep neural networks were first reported in \cite{Szegedy2014ICLR,GoodfellowARXIV2014} demonstrating that deep neural networks are being vulnerable to small adversarial perturbations. This phenomenon was analyzed in several studies \cite{arnab_cvpr_2018,foolai,Fawzi2018,ShahamARXIV2015,FawziNIPS2016}, and different approaches have been proposed to improve the robustness of neural networks \cite{papernot2016distillation,CisseICML2017}. 
Fast Gradient Sign Method (FGSM) and several variations of it were introduced in \cite{GoodfellowARXIV2014,Moosavi-Dezfooli_2016_CVPR} for generating adversarial examples that are indistinguishable--to the human eye--from the original image, but can fool the networks. 
However, these techniques do not apply to state of the art body shape estimation as those are based on multi-stage processing. Typically, shape inference consists in fitting a body model to detected skeleton keypoints. Consequently, we perturb the 2D keypoints to produce an error in the shape fitting step. 
Cisse et al. \cite{CisseNIPS2017}, proposed a method to fool 2D pose estimation. None of these techniques propose a solution to evade model based shape estimation.
In order to evade 3D shape estimation in a subtle manner, we attack by removing and flipping individual keypoints. Since these attacks simulate typical failure modes of detectors (mis-detections due to occlusion and keypoint flips), they are more difficult to identify by the defender.

%% file: userstudy.tex
\section{Understanding Users Shape Privacy Preferences}
Modern body shape methods \cite{sattar19wacv,smplify,omran2018NBF,Kanazawa_2018_CVPR} infer a realistic looking 3D body shape from a single photo of a person. The estimated 3D body captures user weight group and body shape type. However, such a realistic depiction of the undressed body is considered highly private and therefore might not be consented by most people. We performed a user study to explore the users' personal privacy preferences related to their body shape data. Our goal was to study the degree to which various users are sensitive to sharing their shape data such as height, different body part measurement, and their 3D body shape in different contexts. This study was approved by our university's ethical review board and is described next.

\begin{figure}[t!]
  \centering
 \includegraphics[width=0.5\linewidth]{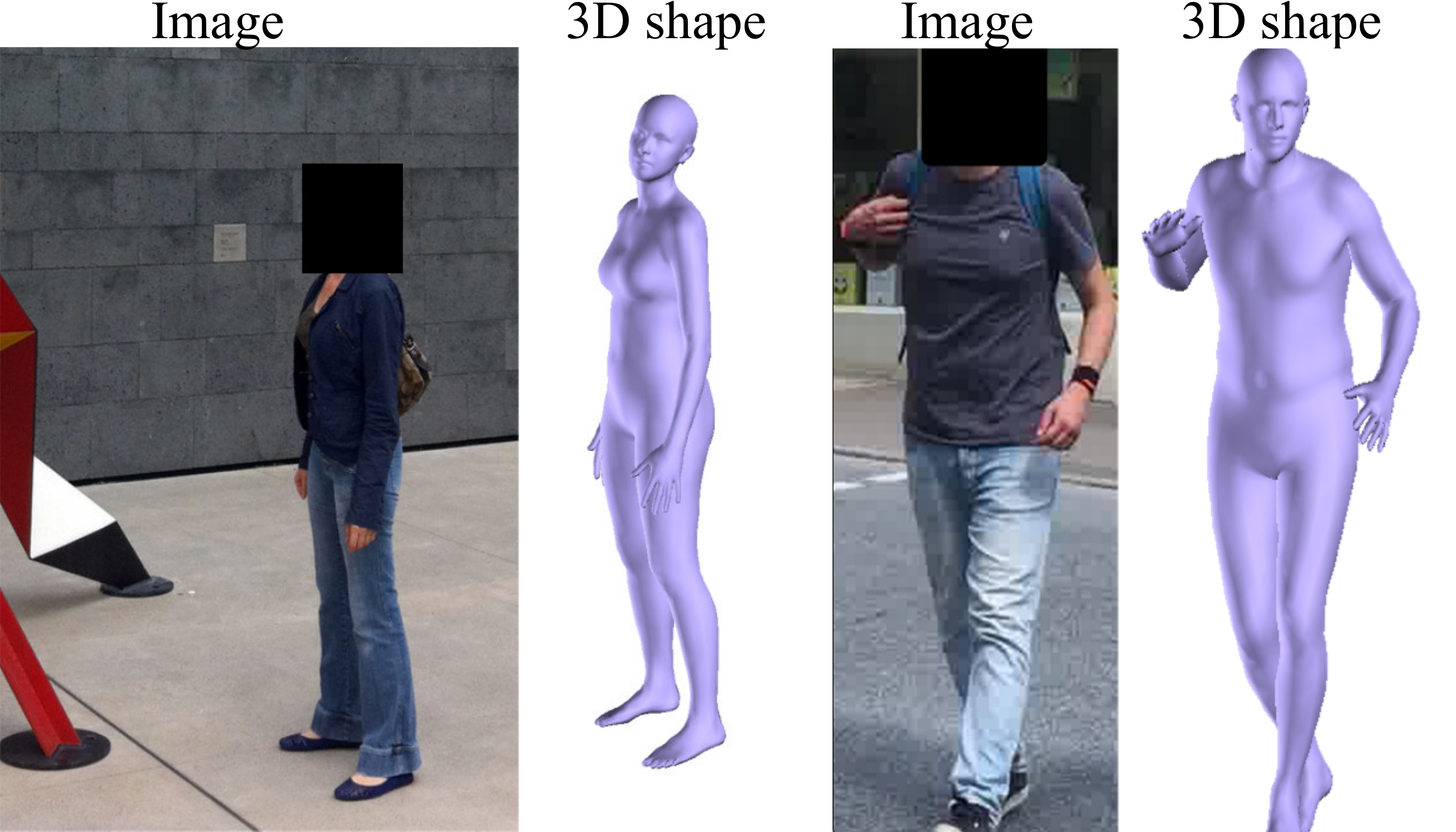}
    \caption{Participants were asked to indicate their comfort level for sharing these images publicly, considering they are the subject in these images.}
    \label{fig:user}
    \vspace{-0.2cm}

\end{figure}
\paragraph{User Study.}
We split the survey into three parts. 
In the first part of the survey, our goal was to understand users image sharing preferences and the users' knowledge of what type of information could be extracted from a single image.

\textit{Part1-Question 1:} Users are shown \autoref{fig:user} without the 3D shape data. Participants are asked how comfortable they are sharing such images publicly, considering they are the subject in these images. Responses are collected on a scale of 1 to 5, where: (1) Extremely comfortable, (2) Slightly comfortable, (3) Somewhat comfortable, (4) Not comfortable, and (5) Extremely uncomfortable. 

\begin{figure}[t!]
  \centering
 \includegraphics[width=0.7\linewidth]{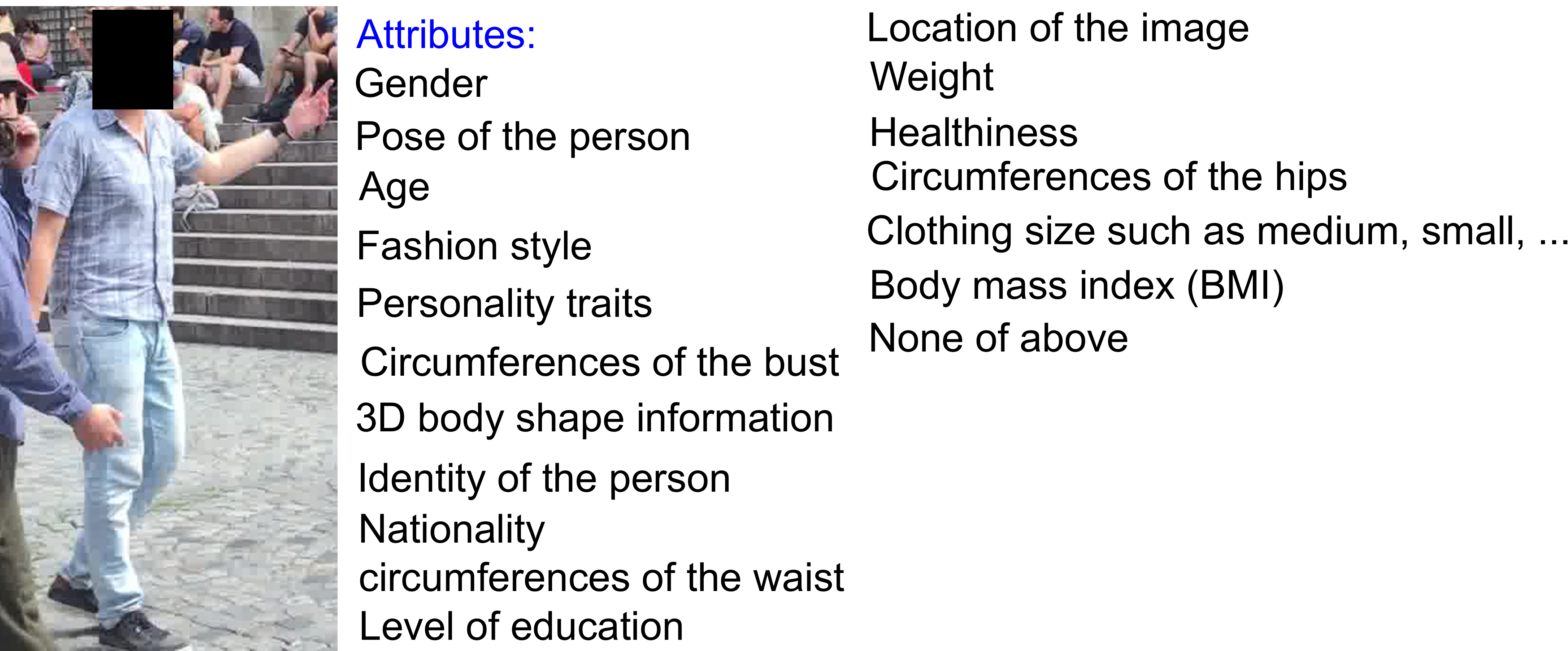}
    \caption{In Question 2 participants were shown this image, and were asked to select the attributes from the list that could be extracted.}
    \label{fig:att}
\end{figure}
\textit{Part1-Question 2:} Participants were shown \autoref{fig:att}, and were asked which attributes could be extracted from this image.

\begin{figure}[t!]
  \centering
 \includegraphics[width=0.6\linewidth]{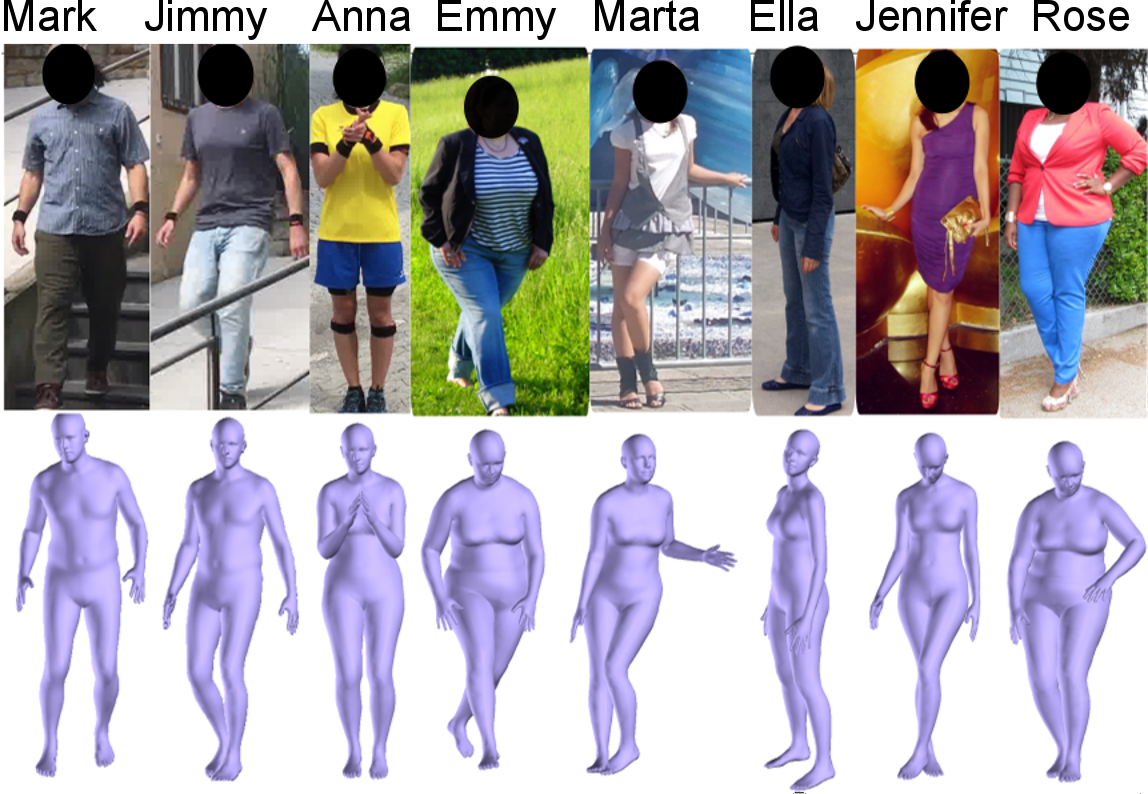}
    \caption{Participants were asked to judge the closeness of the depicted 3D shape to the actual body of the person in the images.}
    \label{fig:close}
    \vspace{-0.5cm}
\end{figure}

In the second part, users were introduced to 3D shape models by showing them images of 8 people along with their 3D body shape, as shown in \autoref{fig:close}. The purpose of part 2 was to understand the user's perceived closeness of extracted 3D shapes to the original images, and their level of comfort with them. 

\textit{Part2-Question 3:}  Participants were asked to rate how close the estimated 3D shape is to the person in the image. Responses are collected on a scale of 1 to 5, where: (1) Untrue of the person in the image, (2) somewhat untrue of the person in the image, (3) Neutral, (4) Somewhat true of the person in the image, and (5) True of the person in the image.  

\textit{Part2-Question 4:} Participants were shown \autoref{fig:user} asked to indicate how comfortable they are sharing such a photograph along with 3D shape data publicly, considering they are the subject in these images. We collected responses on a scale of 1 to 5, similar to Question 1.

In the third part of this survey, we explore users preferences on what type of body shape information they would share for applications such as (a) Health insurance, (b) Body scanners at airport, (c) Online shopping platforms, (d) Dating platforms, and (e) Shape tracking applications (for sport, fitness, ...).

 \textit{Part3-Question 5:} Users were asked their level of comfort on a scale of 1 to 5 for the applications mentioned above.
 
\paragraph{Participants.}
We collect responses of $90$ unique users in this survey. Participants were not paid to take part in this survey. Out of the $90$ respondents, $43.3\%$ were female, $55.6\%$ were male, and $1.1\%$ were queer. The dominant age range of our participants ($63.3\%$) was in 21-39, followed by 30-39 ($23.3\%$). Participants have a wide range of education level, where $46.7\%$  has master degree, $21.1\%$ has bachelor degree\footnote{Further details on participants demographic data are presented in the supplementary materials.}.  
\paragraph{Analysis.}

\begin{figure}[t!]
  \centering
 \includegraphics[width=0.8\linewidth]{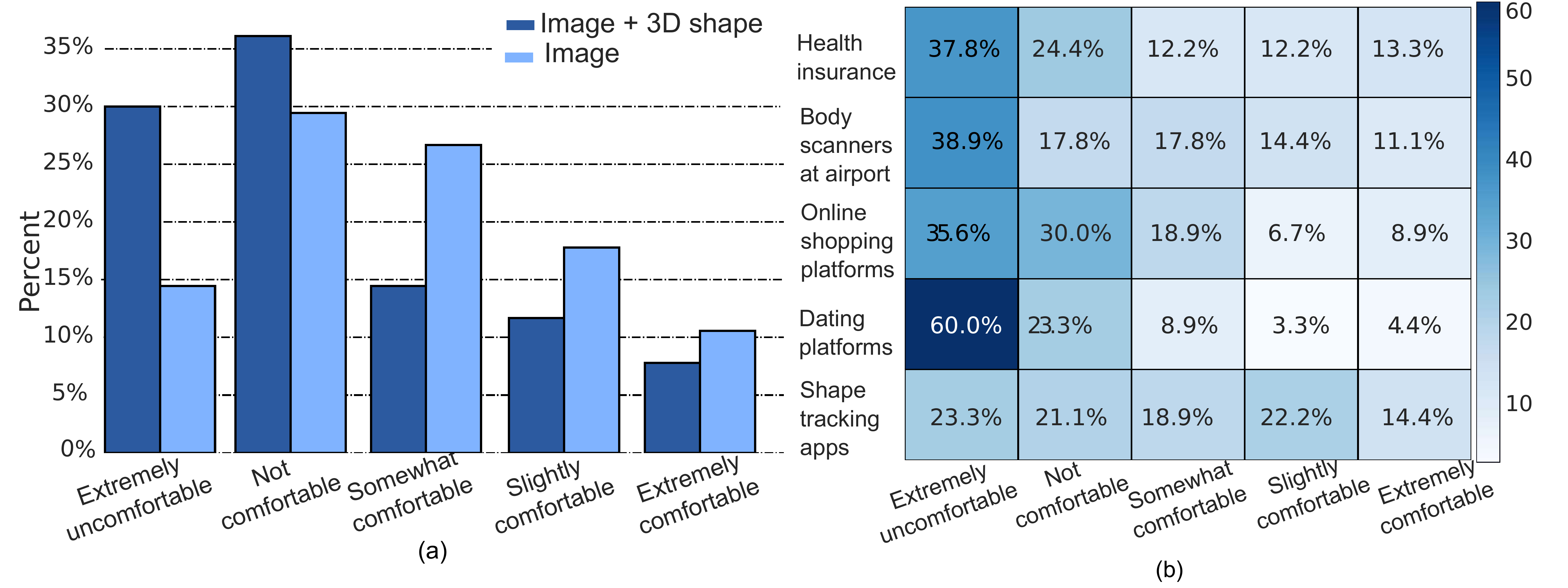}
    \caption{(a) Comfort level of participants in sharing images with and without 3D mesh data, considering they are the subject in these images. (b) Comfort level of the participant for sharing their 3D mesh data with multiple applications. Results are shown as the percentage of times an answer is chosen.}
    \label{fig:app}
\end{figure}

The results of \textit{Part1-Question 1} and \textit{Part2-Question 4} are shown in ~\autoref{fig:app}\textcolor{red}{a}. We see that majority of the users do not feel comfortable or they feel extremely uncomfortable ($ 36.0\%, 30.0\% $) sharing their 3D data publicly compared to sharing only their images ($29.0\%, 14.0\%$).

In \textit{Part1-Question 2}, the top three selected attributes were: 
gender ($98.9\%$), pose ($87.8\%$), and age ($85.6\%$). Shape related attributes such as body mass index (BMI) ($47.8\%$), weight ($63.3\%$), and 3D body shape ($66.7\%$) were not in the top selected attributes, indicating that many participants were unaware that such information could be extracted from an image using automatic techniques.

In \textit{Part3-Question3}, users were asked to judge the quality of the presented 3D models.
Around $43.0\%$ of the participants believe the presented shape is Somewhat true of the person in the image, and $31.0\%$ think the 3D mesh is true to the person in the picture. 
This indicates that recent approaches can infer perceptually faithful 3D body shapes under clothing from a single image.

\autoref{fig:app}\textcolor{red}{b} presents the results from \textit{Part3-Question 5}. Participants show a high level of discomfort in sharing their 3D shape data for multiple applications. In all investigated applications except for fitness, the majority of the users responded with ``discomfort of some degree". 

The user study demonstrates that users are concerned about the privacy of their body shape. Consequently, we present next a framework to prevent 3D shape extraction from images.

%% file: method.tex
\section{Shape Evasion Framework}
\begin{figure*}[t!]
  \centering
 \includegraphics[width=0.9\linewidth]{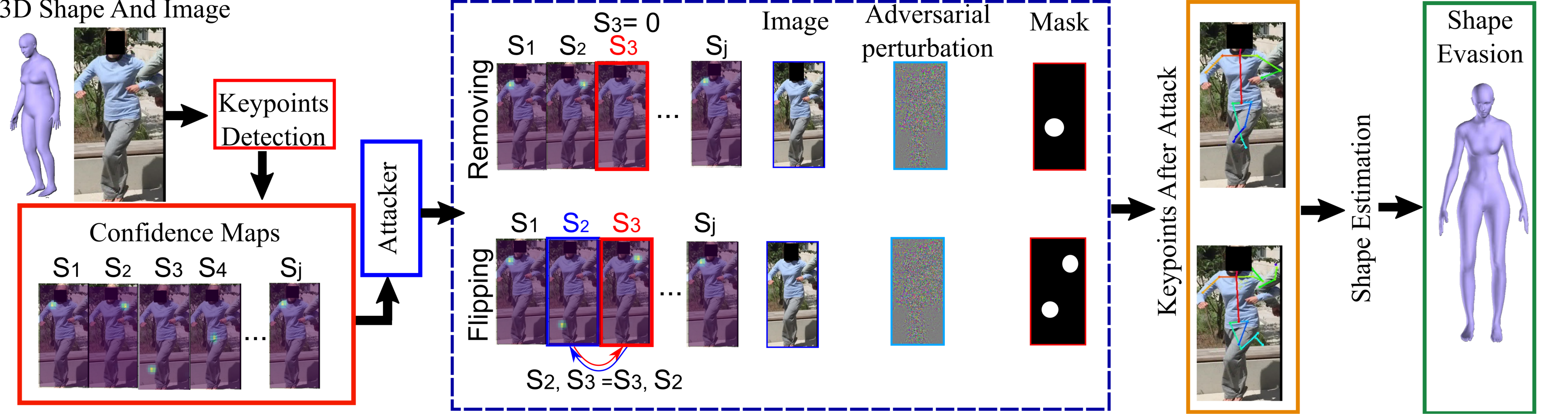}
    \caption{The summary of our framework. We assume that we have full access to the parameters of the network. The attacker breaks the detections by removing or flipping of a keypoint. Hence the final estimated shape does not depict the person in the image.}
    \label{fig:pipline}
   % \vspace{-0.5cm}
\end{figure*}
Model-based shape estimation methods from a 2D image are based on a two-stage approach. First, a neural network is used to detect a set of 2D body keypoints, then a 3D body model fits the detected keypoints. Since this approach is not end-to-end, it does not allows direct computation of the image gradient needed for adversarial perturbation. To this end, we approach the shape evasion by attacking the keypoints detection network. In section~\ref{sec:body}, we give a brief introduction on model-based shape estimation method. In section~\ref{sec:local}, we introduced a local attack that allows targeted attacks on keypoints. \autoref{fig:pipline} shows an overview of our approach. 
\subsection{Model Based Shape Estimation}
\label{sec:body}
The Skinned Multi-Person Linear Model (SMPL)~\cite{SMPL:2015} is a state of the art generative body model. 
The SMPL function $M(\shape, \pose)$, uses shape $\shape$ and poses $\pose$ to parametrize the surface of the human body that is represented using $N = 6890$ vertices. The shape parameters $\shape \in \mathbb{R}^{10}$ encode changes in height, weight and body proportions. The body pose $\pose \in \mathbb{R}^{3P}$, is defined by a skeleton rig with $P = 24$ keypoints. The 3D skeleton keypoints are predicted from body shape via $J(\shape)$. We can use a global rigid transform $R_\theta$ To pose the SMPL keypoint. Hence, $R_\theta(J(\shape)_i)$ denotes a posed 3D keypoint $i$. 
In order to estimate 3D body shape from a 2D image $I$, several works ~\cite{sattar19wacv,smplify,Lassner}, minimize an objective function composed of a keypoint-based data term, pose priors, and a shape prior.
\begin{equation}
E(\shape,\pose)= E_{P_\theta}(\pose)+ E_{P_\beta}(\shape) + E_J(\shape,\pose; \mathbf{K},\mathbf{J}_{\mathrm{est}})
\label{eq:objective}
\end{equation}
where $E_{P_\theta}(\pose)$, and $E_{P_\beta}(\shape)$ are the pose and shape prior terms as described in~\cite{smplify}. The $E_J(\shape,\pose; \mathbf{K},\mathbf{J}_{\mathrm{est}})$ is the keypoint-based data term which penalizes the weighted 2D distance between estimated 2D keypoints,  $\mathbf{J}_{\mathrm{est}}$, and the projected SMPL body keypoint $R_\theta(J(\shape))$: 
\begin{equation}
E_J(\shape,\pose;\mathbf{K},\mathbf{J}_{\mathrm{est}})= \sum_{\mathrm{keypoint} i} w_i\rho(\Pi_{\mathbf{K}}(R_\theta(J(\shape)_i))-\mathbf{J}_{\mathrm{est},i})
\label{eq:single}
\end{equation}
where $\Pi_\mathbf{K}$ is the projection from 3D to 2D of the camera with parameters $\mathbf{K}$ and $\rho$ a Geman-McClure penalty function which is robust to noise in the 2D keypoints detections. $w_i$ indicates the confidence of each keypoints estimate, provided by 2D detection method. For cases such as occluded or missing keypoints, $w$ is very low, and hence the data term will be driven by pose prior term. Furthermore, the prior term avoids impossible poses. 
Shape evasion can be achieved by introducing error in 2D keypoints detection $\textbf{J}_{est}$. We use Adversarial perturbation to fool the pose detection method by either removing a keypoint or filliping two keypoints with each other. 
\subsection{Adversarial Image Generation}
The State-of-the-art 2D pose detection methods such as \cite{cao2018openpose} use a neural network $f$ parametrized by $\phi$, to predict a set of 2D locations of anatomical keypoints $\textbf{J}_{est}$ for each person in the image. The network produces a set of 2D confidence maps $\textbf{S} = \{ \textbf{S}_1,\textbf{S}_2,\textbf{S}_3,...,\textbf{S}_P\}$, where $\textbf{S}_i \in \mathbb{R}^{ w \times h }$ , $i\in {1,2,3,...,P}$, 
is a confidence map for the keypoints $i$ and $P$ is total number of Keypoints. Assuming that a single person is in the image, then each confidence map contains a single peak if the corresponding part is visible. The final set of 2D keypoints $\textbf{J}_{est}$ are achieved by performing non-maximum suppression per each confidence map. These confidence maps are shown in \autoref{fig:pipline}. 

To attack a keypoint we used adversarial perturbation. Adding adversarial perturbation $\textbf{a}$ to an image $\textit{I}$ will causes a neural network to change its prediction \cite{Szegedy2014ICLR}.The adversarial perturbation  $\textbf{a}$ is defined as the solution to the optimization problem

\begin{equation}
\arg \min_{\textbf{a}}  ~\| \textbf{a} \|_2+L(f(I+\textbf{a};\phi),\textbf{S}^*).
\label{eq:adv}
\end{equation}
$L$ is the loss function between the network output and desired  confidence maps $\textbf{S}^*$.

\paragraph{Removing and Flipping of Keypoints:}
The $\textbf{S}^*$ is defined for removing and flipping of keypoints. 
To remove a keypoint, we put its confidence map to zero. For example if we are attacking the first keypoint we have: $ \textbf{S}^*= \{ {\color{red}{\textbf{S}_1 = 0}},\textbf{S}_2,\textbf{S}_3,...,\textbf{S}_P\}$. To flip two key points we exchanged the values of two confidence map as 
$\textbf{S}^i, \textbf{S}^j=\textbf{S}^j, \textbf{S}^i$. In case $i,j=2,3$  we have $
\textbf{S}^*= \{ \textbf{S}_1,\textbf{S}_{\color{red}{3}},\textbf{S}_{\color{red}{2}},...,\textbf{S}_P\}$.
An example of removing and flipping of the keypoint is shown in \autoref{fig:pipline}.
\paragraph{Fast Gradient Sign Method (FGSM) \cite{GoodfellowARXIV2014}:}FGSM is a first order optimization schemes used in practice for \autoref{eq:adv}, which approximately minimizes the $\ell_\infty$ norm of perturbations bounded by the parameter $\epsilon$. The adversarial examples are produced by increasing the loss of the network on the input \textit{I} as
\begin{align}
    I^{\textit{adv}} = I + \epsilon \;  \sign(\nabla_{I} L(f(I;\phi),\textbf{S}^*)).
\end{align}
We call this type of attack global as the perturbation is applied to the whole image. This perturbation results in poses with several missing keypoints or poses outside of natural human pose manifold. While this will often make the subsequent shape optimization step fail (Eq.~\eqref{eq:single}), the approach has two limitations: i) this attack requires a large perturbation and ii) the attack is very easy to identify by the defender. 

\paragraph{Masked Fast Gradient Sign Method (MFGSM):}
\label{sec:local}
To overcome the limitations of the global approach, we introduced Masked FGSM. This allow for localized perturbation for more targeted attacks. This method will generate poses, which are close to ground truth pose, yet have a missing keypoint that will cause shape evasion--while requiring smaller perturbations as shown in the experiments. We will refer to this scheme as ``local'' in the rest of the paper.
To attack a specific keypoint we solve the following optimization problem in a iterative manner as:
\begin{equation*}
I_{0}^{adv} = I 
\end{equation*}
\begin{equation}
I^{\textit{adv}}_{t+1} = \clip(I^{\textit{adv}}_{t} - \alpha \cdot \sign(\nabla_{I^{\textit{adv}}_{t}}L(f(I^{\textit{adv}}_{t}; \phi),\textbf{S}^*) \odot M),\epsilon)
\label{eq:iterative}
 \end{equation}
 where mask $\textbf{M}\in \mathbb{R}^{ w \times h }$ is used to attack a keypoint  $\textbf{J}_{est,i}\in \mathbb{R}^2$ selectively. \textbf{M} is defined as:
\[ \textbf{M} =
  \begin{cases}
   1       & \quad \text{if } (x-\textbf{J}_{est,i})^2 = r^2 \\
   0 &
  \end{cases}
\]
$r$ controls the spread of the attack and $\textbf{x}\in \mathbb{R}^2$ are the pixel coordinates. To ensure the max norm constraint of perturbation $\textbf{a}$ being no greater than $\epsilon$ is preserved, the $clip(z,\epsilon)$ is used, which keeps the values of $z$ in the range$[z-\epsilon, z+\epsilon]$. 

%% file: experiment.tex
\section{Experiments}
\begin{table*}[t]
\centering
\scalebox{.5}{
\begin{tabular}{c|ccccccccccccc|c}
\toprule
Attack & Right ankle & Right knee & Right hip & Left hip & Left knee & Left ankle & Right wrist & Right elbow & Right shoulder & Left shoulder & Left elbow & Left wrist & Head top &Average\\
\midrule
Adversarial & \textbf{1.32} & \textbf{1.4} & 1.39 & 1.37 & \textbf{1.38 }& \textbf{1.32} & \textbf{1.36} & \textbf{1.41} & 1.40 & 1.35 & \textbf{1.28} & \textbf{1.37} & \textbf{1.35} & \textbf{1.37}\\
Synthetic & 1.17 & 1.18 & \textbf{1.79} & \textbf{1.94 }& 1.18 & 1.18 & 1.18 & 1.17 & \textbf{1.43} & \textbf{1.49} & 1.15 & 1.16 & 1.19 & 1.32\\
\bottomrule
\end{tabular}}
\vspace{0.5cm}
\caption{Shape estimation error on 3DPW with Procrustes
analysis with respect to the ground truth shape. Error in cm. The minimum estimation error is 1.16 cm when we have no manipulation on input keypoints. The goal of each attack is to induce bigger error in the estimated shape. Hence, higher errors are indication of a successful attack.}
\label{tab:rem}
\end{table*}

The overall goal of the experimental section is to provide an understanding and the first practical approach to evade body shape estimation and hence protect the privacy of the user. We approach this by systematically studying the effect of attacking keypoint detections on the overall shape estimation pipeline. 
First, we study synthetic attacks based on direct manipulation of keypoints locations, where we can observe the effects on body shape estimation in an idealized scenario. This study is complemented by adversarial image-based attacks which make keypoint estimation fail. Together, we evaluate our approach that provides the first and effective defence against body shape estimation on real-world data.
\paragraph{Dataset.} We used 3D Poses in the Wild Dataset (3DPW) \cite{vonMarcard2018}, which includes 60 sequences with 7 actors. To achieve ground truth shape parameter $\shape$, actors were scanned and SMPL was non-rigidly fit to them to obtain their 3D models similar to \cite{ponsmollSIGGRAPH17clothcap,shape_under_cloth:CVPR17}. To the best of our knowledge, 3DPW is the only in wild image dataset which provides the ground truth shape data as well, which makes this dataset most suitable for our evaluation.
For our evaluation, for each actor, we randomly selected multiple frames from different sequences. All reported results are averaged across subjects and sampled sequence frames.

\paragraph{Model.} We used OpenPose~\cite{cao2017realtime} for keypoint detection as it is the most widely used. OpenPose consists of a two-branch multi-stage CNN, which process images at multi-scales. Each stage in the first branch predicts the confidence map $S$, and each stage in the second branch predicts the Part Affinity Fields (PAFs). For the shape estimation, we used the public code of Smplify \cite{smplify}, which infers a 3D mesh by fitting the SMPL body model to a set of 2D keypoints. To improve the 3D accuracy, we refined the estimations using silhouette as described in \cite{sattar19wacv}. We used MFGSM (Eq. ~\eqref{eq:iterative} with  $\alpha = 1$) in an iterative manner. We evaluated attacks when setting the $\ell_{\infty}$ norm of the perturbations to $\epsilon=0.035$ since we observed that higher values lead to noticeable artifacts in the image. We stop the iterations if we reach an Euclidean distance (between the original and perturbed images) of $0.02$ in image space for local, and $0.04$ for global attacks.

\subsection{Synthetic Modification of The Keypoints}
\label{sec:syn}
First, we studied the importance of each keypoint on the overall body shape estimation by removing one keypoint at a time--which simulates miss-detections.  
The error on shape estimation caused by this attack is reported in the second row of \autoref{tab:rem}. 
We observe that removing ``Hips'', and ``Shoulder'' keypoints results in the highest increase of error of $60.78\%$, and $25.86\%$ whereas ``Elbows" and ``Wrists" result in an increase of only $1\%$.

\begin{figure}[!h]
\centering
\includegraphics[width=0.95\linewidth]{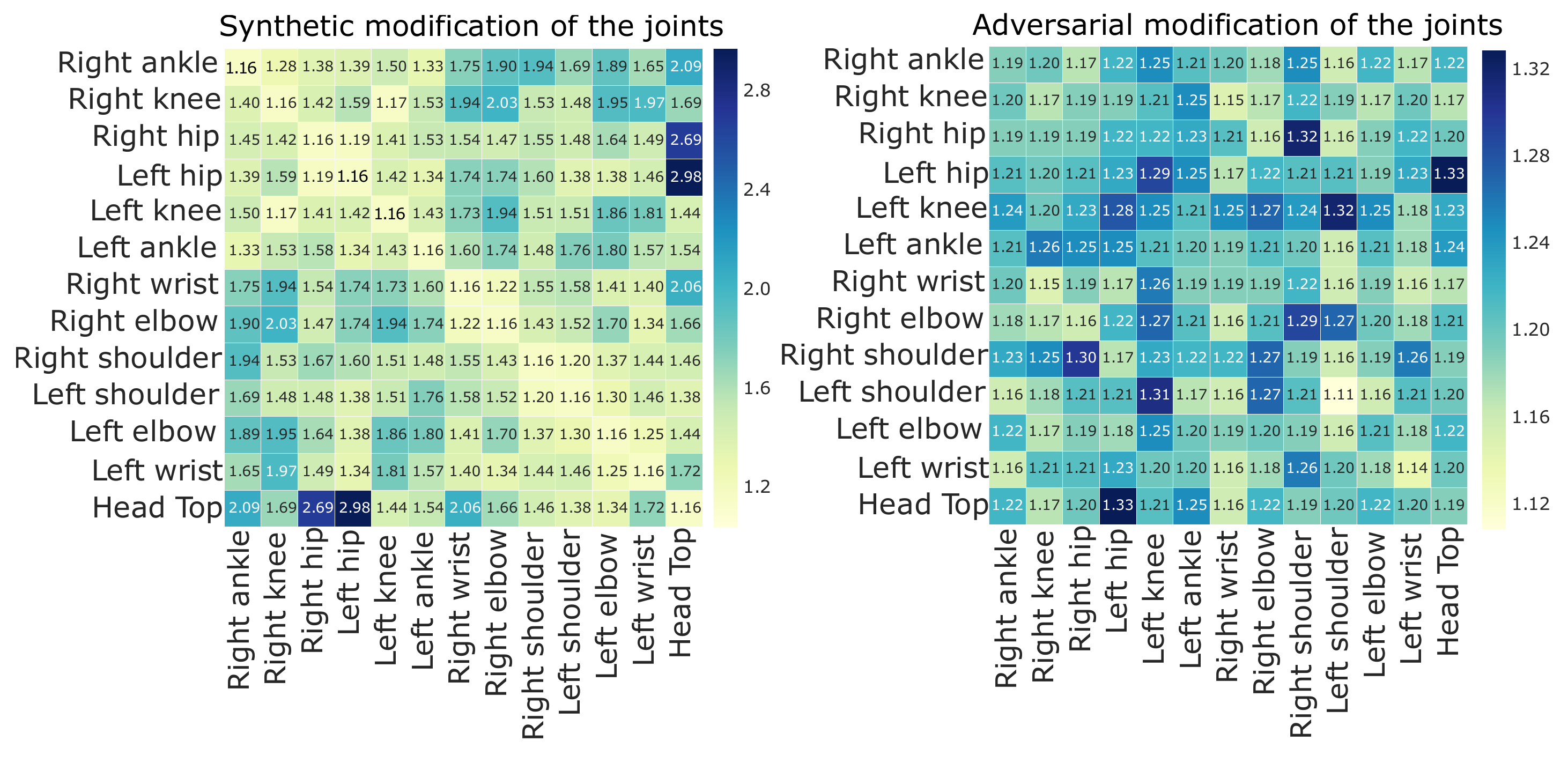}
\caption{Shape estimation error on 3DPW with Procrustes analysis. Error in cm for synthetic and adversarial flipping of the keypoints.}
\label{fig:flip}
%\vspace{-0.5cm}
\end{figure}

\begin{figure}[!t]
\centering
\includegraphics[width=0.5\linewidth]{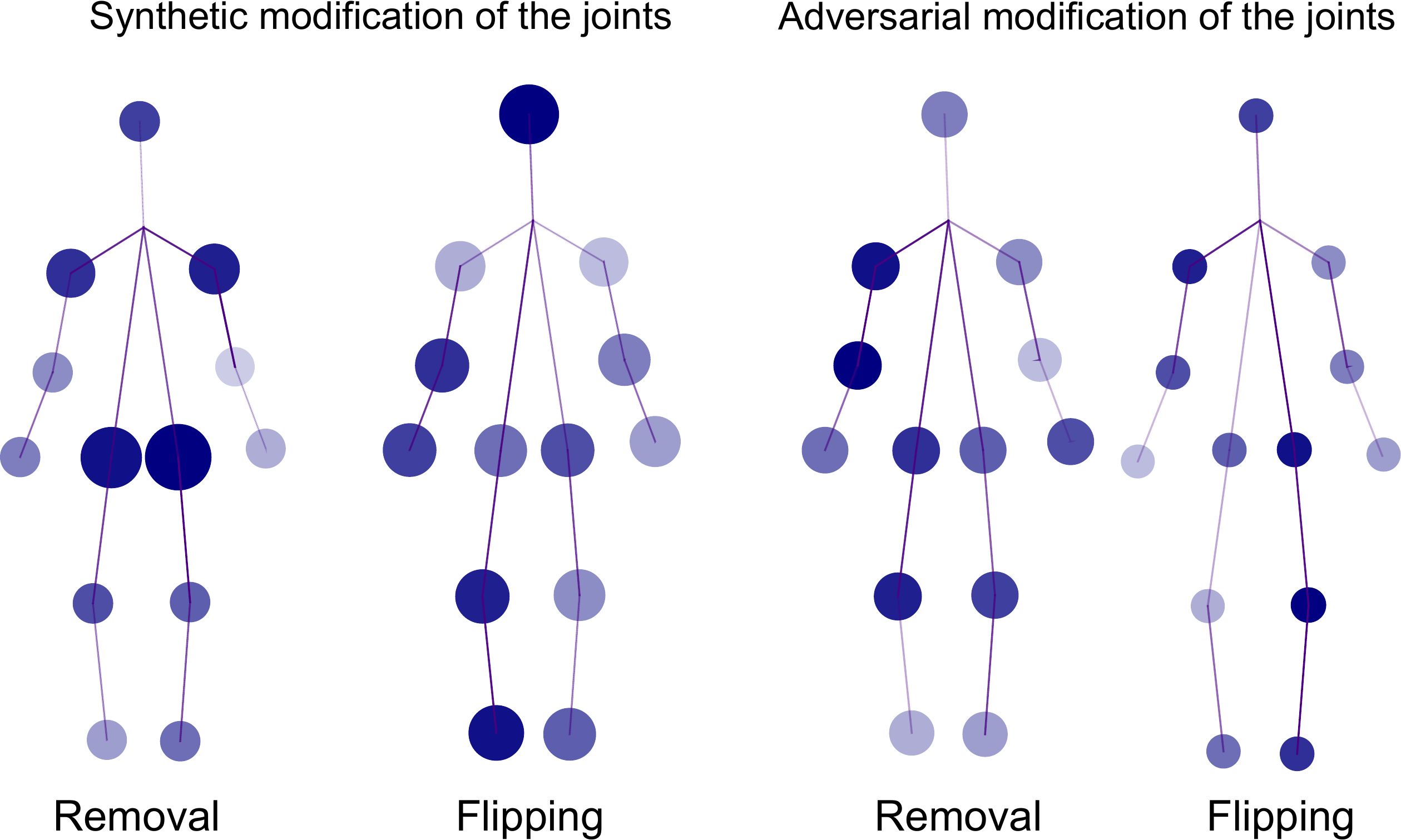}
\caption{The overall shape estimation error induced by synthetic and adversarial (local) attacks. The darker and bigger circles shows higher error.}
\label{fig:stick}
\end{figure}

We also studied the effect of flipping keypoints. The results of this experiment are shown in \autoref{fig:flip}. Flipping the ``Head'' with the left or right ``Hip'' caused an increase in error of $143.96\%$. Flipping the ``Elbow'' and ``Knee'' was the second most effective attack causing $67.0\%$ increase of error in average. The least effective attack was by flipping the left and right knee (less than $0.1\%$). 
The average error introduced by removing or flipping of each keypoint is illustrated in \autoref{fig:stick} -- higher error is larger in size and darker in colour. We can see that, overall ``Hip'', ``Shoulder'', and ``Head'' keypoints play a crucial role in the quality of the final estimated 3D mesh, and are the most powerful attacks. 

\subsection{Attacking keypoint Detection by Adversarial Image Perturbation}
\begin{figure}[t]
\centering
\includegraphics[width=0.9\linewidth, height=\linewidth,keepaspectratio]{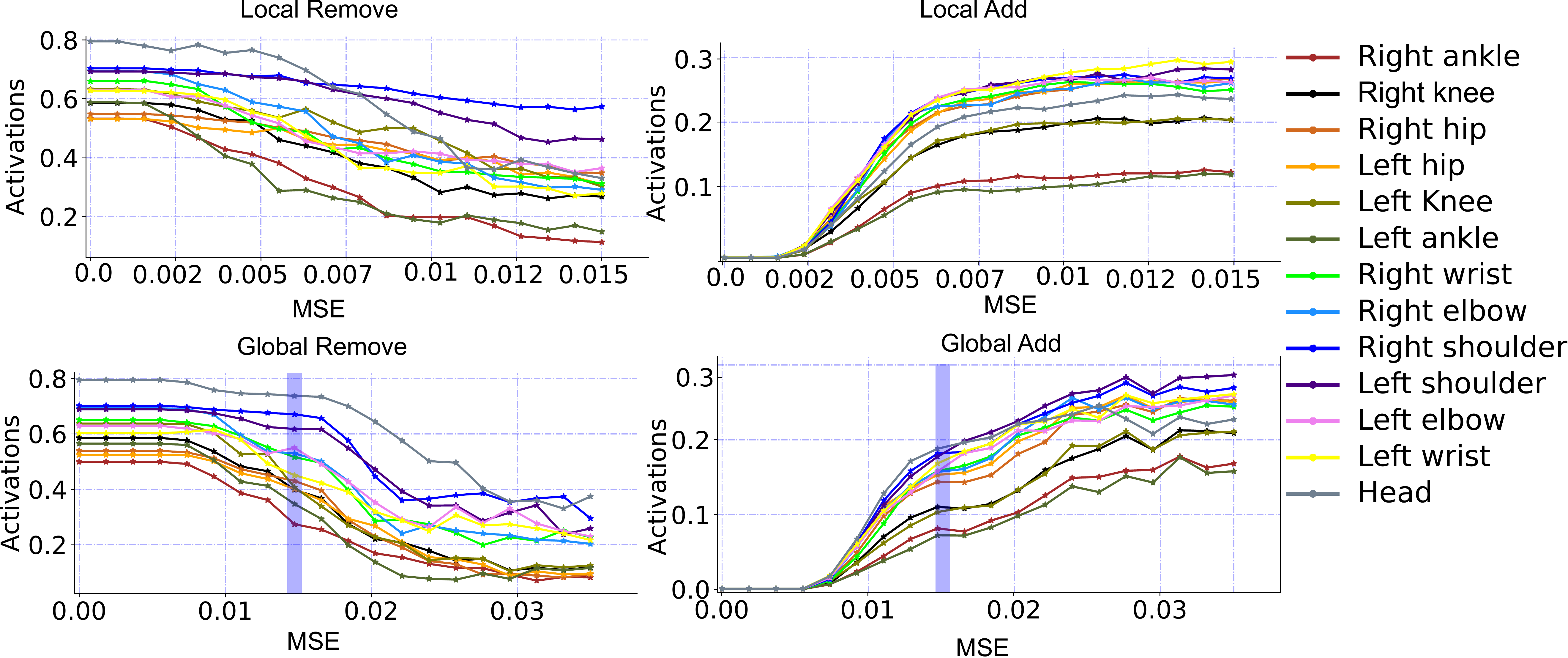}
\caption{Comparison of local and global attacks for removing and adding a keypoint. The local attack has a higher rate of decrease or increase of activation compared to the global method for the same amount of perturbation. The blue bar on the global plots shows the end of the local methods.}
\label{fig:act}
\vspace{-0.5cm}
\end{figure}

To apply modifications to the keypoints, we used our proposed local Mask Iterative Fast Gradient Sign Method (MIFGSM). 
\autoref{fig:act} shows the keypoint confidence map values when removing and adding a keypoint using local and global attacks with respect to the amount of perturbation added to the image. 
We can see that the activation per keypoint decreases after each iteration. Interestingly, the rate of decrease is slower for global attacks for the same amount of perturbation ($0.015$ Mean Squared Error (MSE) between perturbed and original image). Global attacks require a much higher amount of perturbations ($0.035$ MSE) to be successful, causing visible artifacts in the image. We observed similar behavior when adding ``fake" keypoint detections (required to flip two keypoints). Similarly, the rate of increase in activation was slower for global compared to local for the same amount of perturbation ($0.015$ MSE, blue bar in the plot).
From \autoref{fig:act} we can also see that shoulders and head are more resistant to the removal. Furthermore, the attack was the most successful in the creation of wrists. Since local attacks are more effective, we consider only the local attack method for further analysis.

\subsection{Shape Evasion}
In this section, we evaluate the effectiveness of the whole approach for evading shape estimation and therefore, protecting the users' privacy. We used our proposed local method to remove and flip keypoints instead of the synthetic modification of the keypoints as described in \textcolor{red}{section} \ref{sec:syn}. Hence, we call this attack as a adversarial modification of keypoints. 

The error on shape estimation caused by removing of the keypoints using our local method are reported in the first row of \autoref{tab:rem}, we refer to it as adversarial. We see that attacks on ``Right Elbow'' and ``Right knee'' causes $21.55\%$, and $20.69\%$ increase of error in shape estimation. The least amount of error $10.34\%$ and $13.79\%$ was produced by removing ``Left Elbow'' and ``Ankles'' respectively. However, ``Hip'' and ``Shoulder'' gained higher error in average for left and right keypoint by $19.0\%$. On average, the adversarial attack for removing keypoints caused an even higher error than the synthetic mode ($18.1\%$ to $13.79\%$), showing the effectiveness of this approach in shape evasion and hence protecting the users' privacy. 

The result for flipping the keypoints is shown in \autoref{fig:flip} (Adversarial modification of the keypoint). 
The highest increase in error was ($14.66\%$) caused by flipping the ``Head'' with ``Left Hip'', the second most effective attack was for flipping the ``Right Shoulder'' and ``Left Knee'' keypoints ($12.93\%$). Overall the most effective attack was on ``Knee'' and the least effective attack was on ``Wrist'' with increase of $7.11\%$, and $1.72\%$ error on average, respectively. 

Adversarial flipping of keypoints achieves an error of $4.19\%$ compared to adversarial removing attacks ($18.1\%$), which shows they are less effective. In addition, similar to global attacks, filliping of keypoints causes more changes in the keypoints, making the detection of these attacks easier.
\begin{figure}
    \centering
    \begin{subfigure}[b]{0.49\linewidth}
        \includegraphics[width=\linewidth]{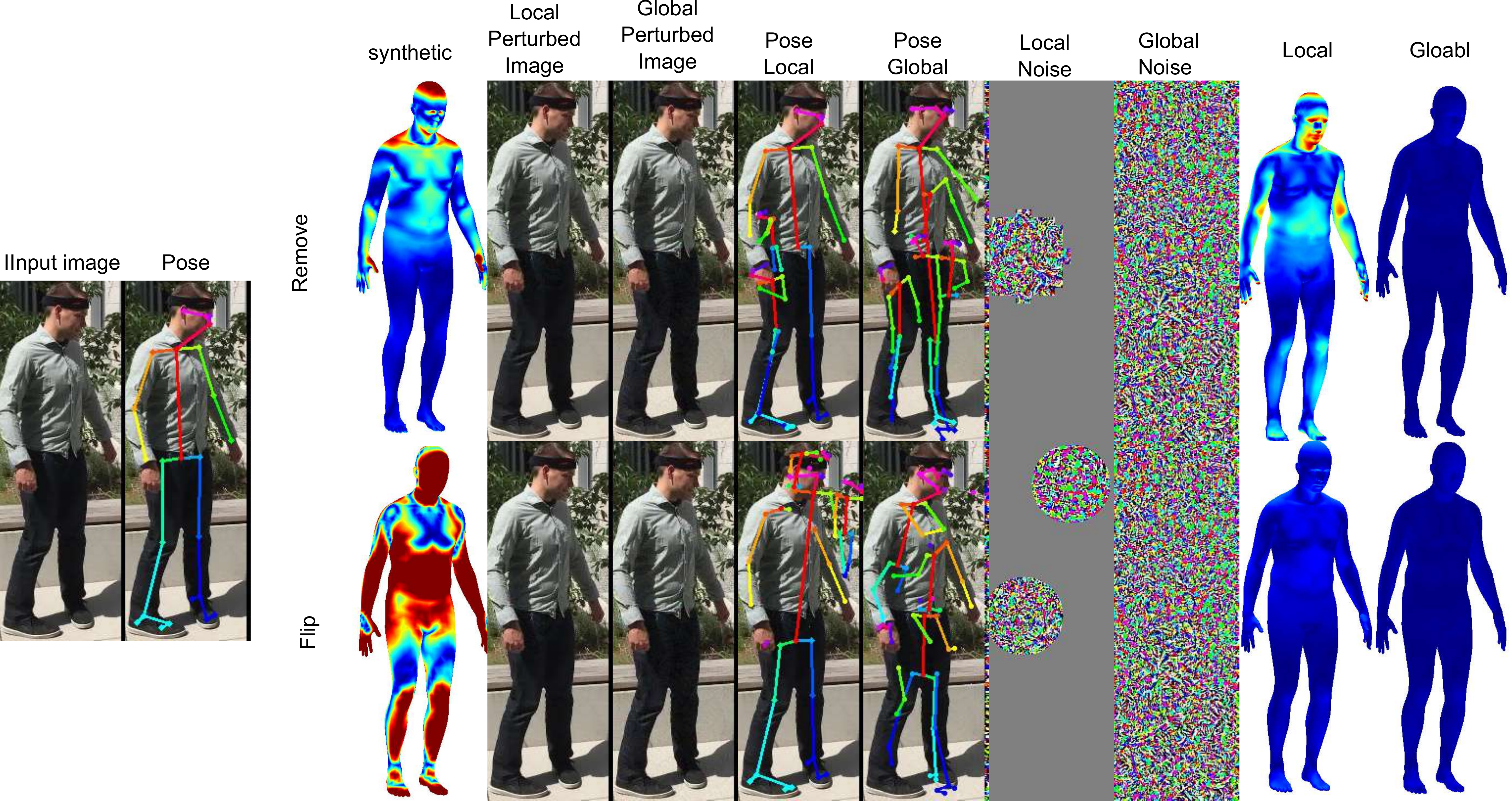}
        \caption{Person with body shape close to SMPL template ($0.04$ cm)}
        \label{fig:exp1}
    \end{subfigure}
    ~ 
    \begin{subfigure}[b]{0.4\linewidth}
        \includegraphics[width=\linewidth]{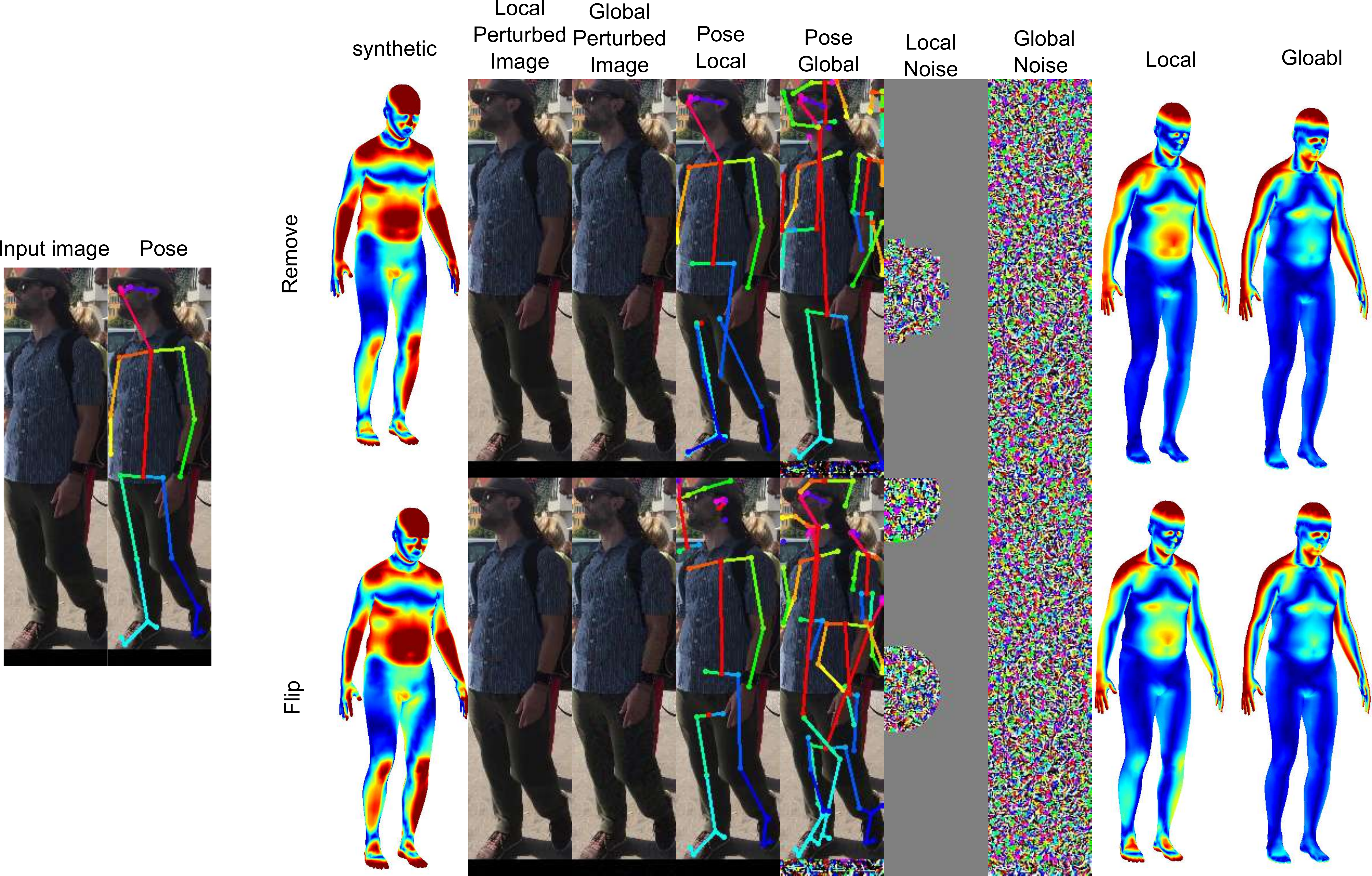}
        \caption{Person with a higher distance to SMPL template ($2.0$ cm)}
        \label{fig:exp2}
    \end{subfigure}
\caption{The left side shows the original image with the estimated pose, and the right the output when modified with local and global adversarial perturbations with corresponding error heatmaps with respect to ground truth shapes (red means $>$ $2.0$ cm). Here we applied local and global attack for removing the ``Right Hip'', and flipping the ``Right Hip'' and ``Head Top''. The global attack causes the pose estimation to hallucinate multiple people in the image, while our local attack only changes the selected keypoints. The predicted shape in case of a global attack is always close to the average template of SMPL causing a lower error for people with an average shape.}\label{fig:quantitavie}
\vspace{-0.5cm}
\end{figure}

\subsection{Qualitative Results}
In \autoref{fig:quantitavie}, we present example results obtained for each type of attack. 
The global attack causes pose estimation to hallucinate multiple people in the image, destroying the body signal of the person in the picture. As the predicted poses in the global attack are not in human body manifold, the optimization step in SMPL will fail to fit these keypoints resulting in average shape estimates. 
 In the local attack, we were able to apply small changes in the keypoints. Hence, these small changes make the shape optimization stage predict shapes that are not average and also not close to the person in the image. Overall, shape evasion was most successful when removing the keypoints than flipping them, and when using the local attacks.

\section{Discussion}
As our study of privacy on automatically extracted body shapes and method for evading shape estimation is the first of its kind, it serves as a starting point -- but naturally needs further investigations to extend on both lines of research that we have touched on. The following presents a selection of open research questions.
\vspace{-0.2cm}     
\paragraph{Targeted vs untargeted shape evasion.} While our method for influencing the keypoints detection is targeted, the overall approach to shape evasion remains untargeted. Depending on the application scenario, a consistent change or particular randomization of the change in shape might be desired, which is not addressed by our work.
\vspace{-0.2cm}     
\paragraph{Effects of adversarial training.} It is well known that adversarial training against particular image perturbations can lead to some robustness against such attacks \cite{Szegedy2014ICLR,oh17iccv} and in turn, the attack can again be made to some extent robust against such defences. Preventing this cat-mouse-game is subject of on-going research and -- while very important -- we consider outside of the scope of our first demonstration of shape evasion methods. 
\vspace{-0.2cm}     
\paragraph{Scope of the user study.} While our user study encompasses essential aspects of privacy of body shape information, clearly a more detailed understanding can be helpful to inform the design evasion techniques and privacy-preserving methodologies that comply with the users' expectations on handling personal data. As our study shows that such privacy preferences are personal as well as application domain specific, there seem ample opportunities to leverage the emerging methods of high-quality body shape estimation in compliance with user privacy.

\section{Conclusion}
Methods for body shape estimation from images of clothed people are getting more and more accurate. Hence we have asked the timely question to what extent this raises privacy concerns and if there are ways to evade shape estimation from images.
To better understand the privacy concerns, we conduct a user study that sheds light on the privacy implication as well as the sensitivity of shape data in different application scenarios. Overall, we observe a high sensitivity, which is also dependent on the use case of the data. Based on this understanding, we follow up with a defence mechanism that can hamper or even prevent body shape estimation from real-world images. Today's state of the art body shape estimation approaches are frequently optimization based and therefore don't lend themselves to gradient-based adversarial perturbation. We tackle this problem by a two-stage approach that first analysis the effect of individual keypoints on the shape estimate and then proposes adversarial image perturbations to influence the keypoints. In particular, our novel localized perturbation techniques constitute an effective technique to evade body shape estimation at negligible changes to the original image.

%% file: supp.tex
\section{Body Shape Privacy in Images: Understanding Privacy and Preventing Automatic Shape Extraction\\ Supplementary  Material}
We are providing here more details about the survey on exploring the users' personal privacy preferences related to their body shape data. 
This user study was split into three sections: 1) Usage of online platforms, 2) 3D shape models, and 3) application of 3D shape models. 
\subsection{Usage of online platforms:}
\textbf{\textit{Question 1:}}
In this section, we would like to get more information about your usage of online platforms. 
On an online public platform, you create an account. On this platform, you are allowed to post photographs, which anyone can view. Moreover, you can also interact with other users who shared their photographs and can comment on or like them.

Imagine you are the person in these photographs, what's your comfort level in posting them online? \\

{\centering
\includegraphics[width=0.35\linewidth]{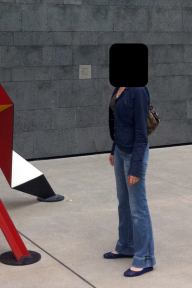}
\includegraphics[width=0.195\linewidth]{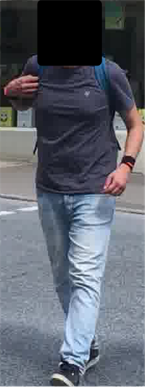}

}
On a scale of 1-5, rate how comfortable you are sharing such photographs, where:
\begin{enumerate}
\item You are extremely comfortable sharing such data
\item You are slightly comfortable sharing such data
\item You are somewhat comfortable sharing such data
\item You are not comfortable sharing such data
\item You are extremely uncomfortable sharing such data
\end{enumerate}

\begin{figure}[t]
    \centering
    \includegraphics[width=\linewidth,height=\linewidth, keepaspectratio]{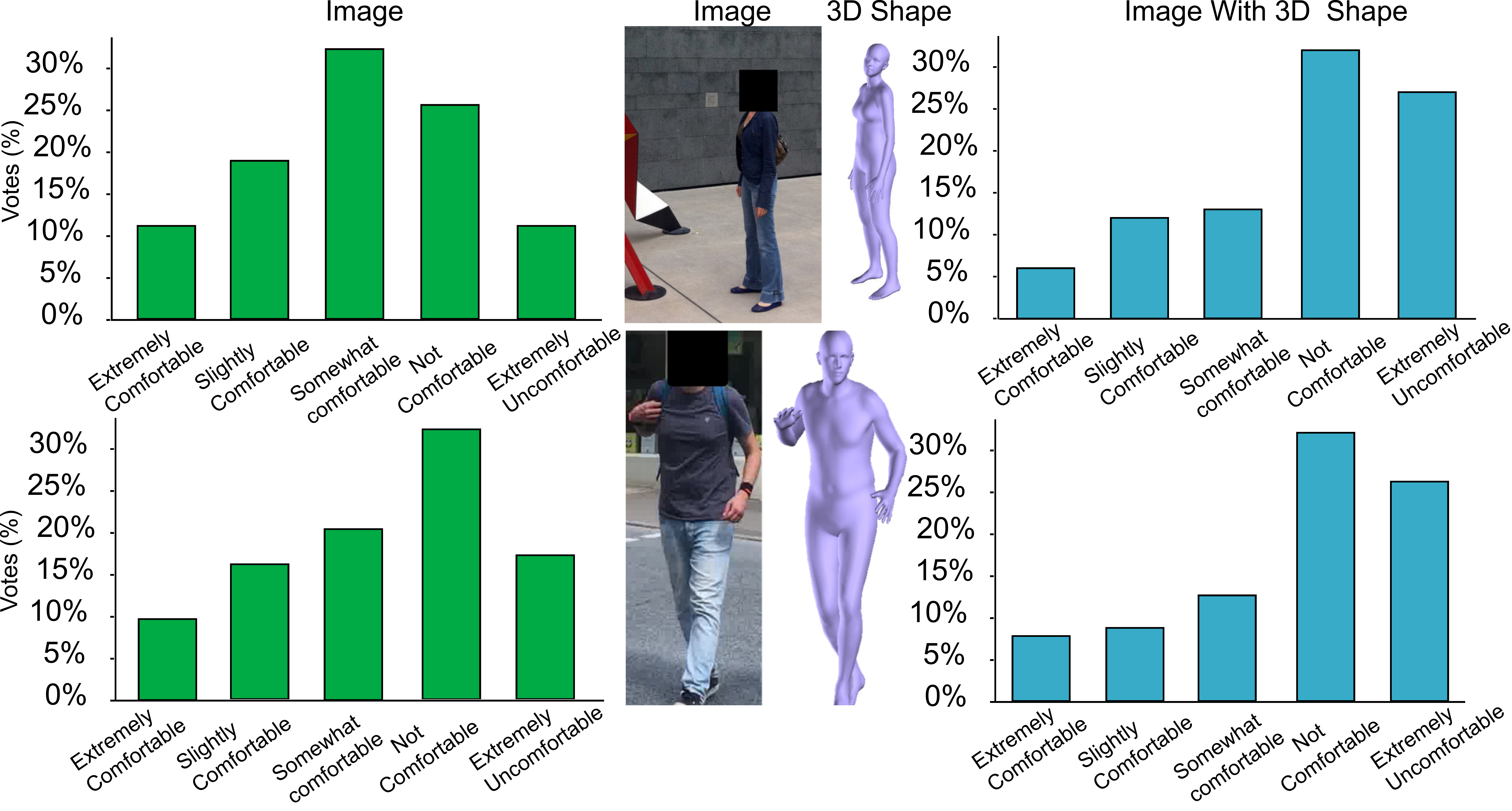}
    \caption{Participants were shown the image with and without 3D shape data and were asked to indicate their comfort level for sharing this data publicly. One can see that the majority of participant have a high level of discomfort in sharing their 3D shape data publicly. This can be seen especially for the  female subject were the comfort distribution towed towards not comfortable when having 3D shape data along.}
    \label{fig:3D}
\end{figure}

\textbf{\textit{Answer:}} The result of this question is shown in \autoref{fig:3D}. Participants were shown the figure in the middle. The figure on left shows the users' comfort level in sharing each image without the 3D shape data online. 

\textbf{\textit{Question2 :}}
Have you ever uploaded your image to online shopping sites, social media, or fashion blogs?

\textbf{\textit{Answer:}} Around $68.5\%$ of our participant answered yes to this question.

\textbf{\textit{Question 3:}}
What information could be extracted from this image?

{\centering
\includegraphics[width=0.2\linewidth]{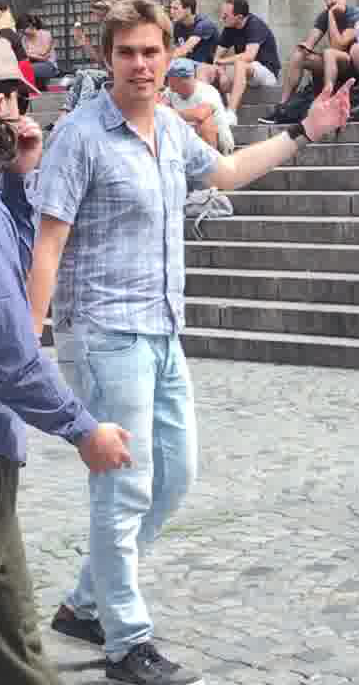}

}
\begin{figure}[t]
    \centering
    \includegraphics[width=\linewidth,height=\linewidth, keepaspectratio]{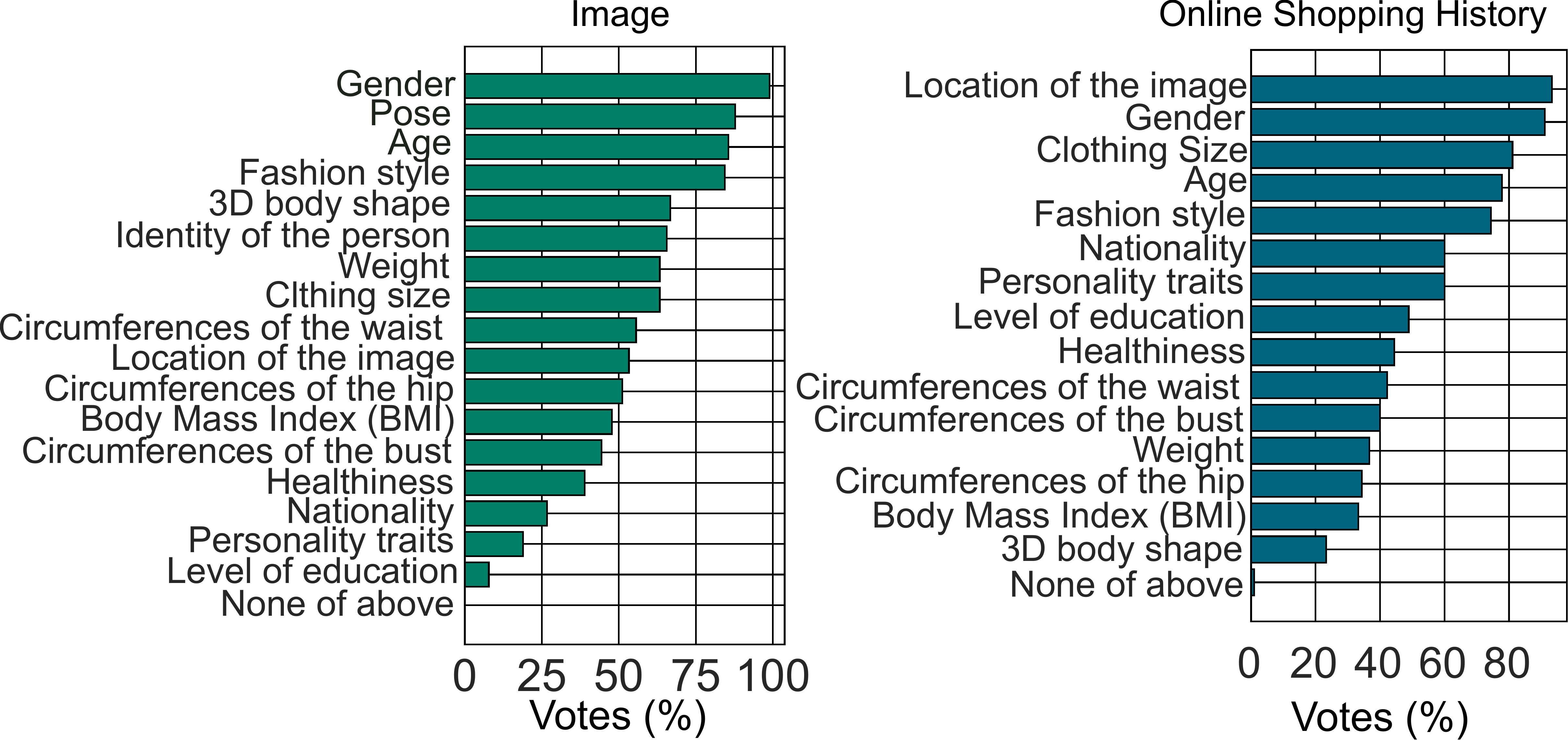}
    \caption{List of attributes selected by our participants in \textbf{\textit{Question 3}}, and \textbf{\textit{Question 4}} }
    \label{fig:history}
\end{figure}

\textbf{\textit{Question 4:}}
What type of information could online shopping platforms (e.g. Amazon) extract from your purchase history?

\textbf{\textit{Answer:}} The answer to  \textit{Question3}, and \textit{Question 4} are shown in \autoref{fig:history}. The list is ordered with the number of votes. We can see that the shape related attributes were not in the top selected ones, showing a lack of awareness of participants on recent technologies for shape extraction from a single image. Some of non-relevant attributes were removed from Question 5.

\subsection{3D Shape Models}
In this section, we introduced our participant with 3D shape data with the following text.

\textbf{\textit{Question 5:}} Recent advances in computer vision make it possible to extract 3D shape data from a single image. In this figure, you can see several examples of the estimated shape of a person from their image.
Please tick one box to rate how well the depicted 3D shapes reflect the person in this figure.

{\centering
\includegraphics[width=\linewidth]{figure/drawing-1.png}

}
The answer was collected in scale 1 to 5 as:
\begin{enumerate}
    \item Untrue of the person in the image
    \item Somewhat untrue of the person in the image
    \item Neutral
    \item Somewhat true of the person in the image
    \item True of the person in the image
\end{enumerate}
\begin{figure}[t]
    \includegraphics[width=\linewidth,height=\linewidth, keepaspectratio]{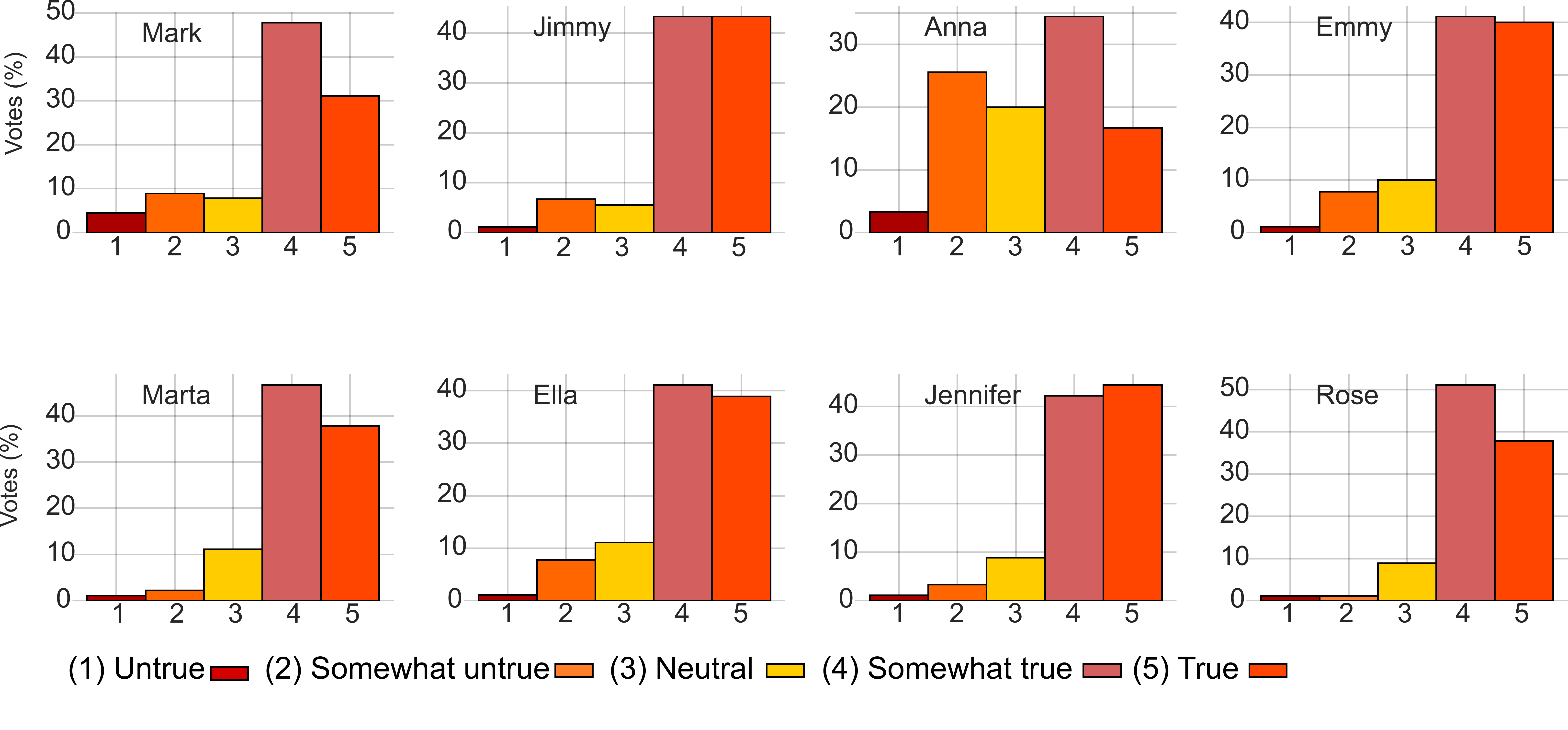}
    \caption{Participant were asked to rate how well the depicted 3D shape reflects the person in the image. Somewhat true and true was selected by majority of participants.}
    \label{fig:b}
\end{figure}

\textbf{\textit{Answer:}} The answer to this question is depicted in \autoref{fig:b}. The majority of our participants think the 3D shapes are true estimations of the person shape. 

\textbf{\textit{Question 6:}}
Imagine you are the person in these photographs \autoref{fig:3D}, what's your comfort level in posting these photograph along with the 3D shape data online?

{\centering
\includegraphics[width=0.32\linewidth]{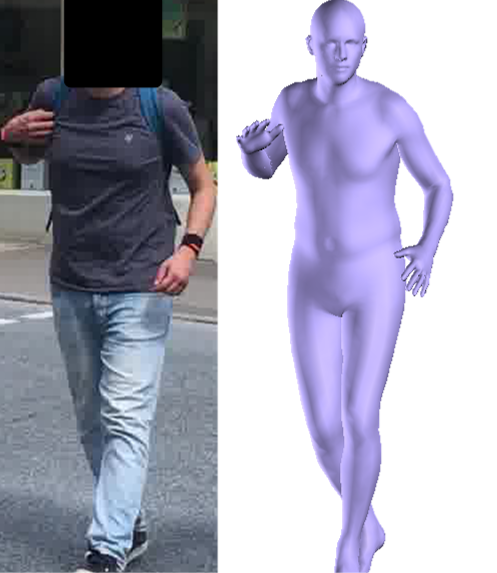}
\includegraphics[width=0.4\linewidth]{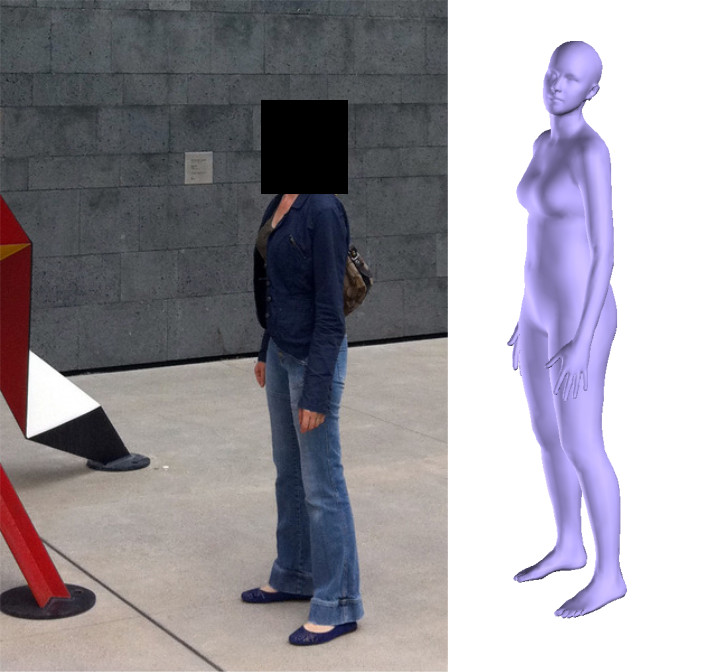}

}

\vspace{1cm}
\textbf{\textit{Answer:}} The answer to this question is depicted in \autoref{fig:3D}. We can see that people change their sharing preferences when the 3D shape is available. This can be especially seen for the female subject. 
\subsection{Application of 3D Shape Models}
3D shape data can be acquired and use in different applications. For example, the 3D scanner at the airport could have exact detailed body model of the person. Online stores could use their customer's 3D shape data in their recommendations systems. These data could be also used to capture your progress when you go to the gym.  Your shape information could be used by the insurance company to get an estimation of your health, or dating web pages could provide this information for other users and you to improve their match predictions.  

\textbf{\textit{Question 7:}}
what is your comfort level in sharing your 3D shape data with these applications?

\textbf{\textit{Answer:}} The answer to this question was presented in \textcolor{red}{Figure 5.b} in the main paper. Participants show a high level of discomfort in sharing their 3D shape data with multiple applications. In all investigated applications except fitness, the majority of users express some degree of discomfort.

\subsection{Demographic information }
We collected three types of demographic data.
Which category below includes your age? What is your gender?, and What is the highest level of school you have completed or the highest degree you have received?

\textbf{\textit{Answer:}} The summary of this data is given in \autoref{tab:demo}.

\begin{table}[h]
\centering
\scalebox{.95}{
\begin{tabular}{lccc}
\toprule
\textbf{Gender} & Female & Male & Other \\
& 39& 50& 1\\
\bottomrule
\textbf{Age}  & &\\
18 - 20 & 2 & 2 & 0 \\
21 - 29 & 22 & 34 & 1\\
30 - 39 & 10 & 11 & 0 \\
40 - 49 & 3 & 2 & 0\\
50 - 59 & 1 & 0 & 0\\
60 or older  & 1 & 1 & 0\\
\bottomrule
\textbf{Highest Completed Education} & &\\
High school degree or equivalent (e.g., GED) & 3 & 4 & 0 \\
Some college but no degree & 1 & 0 & 0 \\
Bachelor’s degree & 7 & 11 & 1\\ 
Master’s degree & 20 & 22 & 0\\ 
Graduate degree & 1 & 3 & 0\\ 
Professional degree & 2 & 0 & 0\\
Doctorate degree & 5 & 10 & 0\\
\bottomrule
\end{tabular}}
\caption{\textbf{Participants demographic. Total N = 90}. Majority of our participant were in age range 21-29 with Masters degree.}
\label{tab:demo}
\end{table}
\clearpage